\documentclass[11pt]{article}
\usepackage[margin=1in]{geometry}

\usepackage{times}
\usepackage{latexsym}
\usepackage{subcaption}
\usepackage{booktabs}
\usepackage[T1]{fontenc}

\usepackage[utf8]{inputenc}

\usepackage{microtype}

\usepackage{inconsolata}

\usepackage{graphicx}

\usepackage[numbers]{natbib}
\usepackage[hidelinks]{hyperref}
\usepackage{xurl}

\title{Causal inference and model explainability analysis of a real world sales dataset}
\date{}

\author{
  Pranav Gupta\\
  Cisco\footnote{work done when the author was working for Lowe's Companies Inc} \\
  Seattle, WA, USA \\
  \\\and
  Nithin Surendran \\
  Lowe's Companies Inc \\
  Mooresville, NC, USA \\
  \texttt{nithin.surendran01@lowes.com} \\
  (corresponding)
}

\begin{document}
\maketitle
\begin{abstract}
Most major retailers today have multiple divisions focused on various aspects, such as marketing, supply chain, online customer experience, store customer experience, employee productivity, and vendor fulfillment. They also regularly collect data corresponding to all these aspects as dashboards and weekly/monthly/quarterly reports. Although several machine learning and statistical techniques have been in place to analyze and predict key metrics, such models typically lack interpretability. Moreover, such techniques also do not allow the validation or discovery of causal links. In this paper, we aim to provide a recipe for applying model interpretability and causal inference for deriving sales insights. In this paper, we review the existing literature on causal inference and interpretability in the context of problems in e-commerce and retail, and apply them to a real-world dataset. We find that an inherently explainable model has a lower variance of SHAP values, and show that including multiple confounders through a double machine learning approach allows us to get the correct sign of causal effect. 
\\
\end{abstract}


\section{\label{sec:intro}Introduction}

Large-scale retail enterprises have complex ecosystems that involve marketing, supply chain management, online and in-store customer experiences, employee management, and vendor relations. The interplay among these divisions directly influences the overarching success metric of retail enterprises-- sales. Hence, retail enterprises regularly collect extensive data across these divisions to drive decision-making processes. Executives in such enterprises are not only concerned with the ``factuals,'' i.e., the data for the current reality, but also the ``counterfactuals,''\cite{causal-biz-book, PearlMackenzie18} i.e., what if they invested in marketing? What if they invested more in vendor management?


Retailers frequently use several machine learning models to analyze and predict key performance metrics. However, these models often lack interpretability, making it difficult for stakeholders to understand the underlying mechanisms driving predictions. Furthermore, existing techniques fall short in facilitating the validation or discovery of causal relationships between different operational aspects and sales metrics. Given the complexity of retail enterprises, it is extremely difficult to generate a realistic counterfactual scenario for feeding into a predictive model, even if the model approaches nearly $100\%$ performance during training and evaluation. For example, imagine a CEO whose data science team has built a highly accurate model that predicts sales as a function of warehouse inventory, marketing spend, and several other features. The CEO wants to understand the intuition behind the model's decisions, but her team tells her that the model is a deep neural network, whose hidden layer weights and intermediate hidden states cannot be directly mapped to an intuitive understanding. While model interpretability is less critical in certain e-commerce applications, in this case it is much more critical because of the high stakes associated with each executive-level decision.

There is another major problem here. Let us say that the model shows that increasing the marketing spend by $5\%$ will lead to a $10\%$ increase in sales. However, we do not know what the actual inventory in the ``counterfactual'' scenario of increased marketing spend. Maybe promotions will cause an increase in demand and the demand will reduce the inventory. There could be partial redundancy between inventory and marketing spend, or a confounding aspect that drives both inventory and marketing spend. The CEO needs a highly accurate prediction of the ``counterfactual'', before feeding the counterfactual into her team's predictive model. To predict an accurate ``counterfactual,'' the data science team needs to accurately model the generative process and have all the variables in the generative process accurately documented. Many a time, these ``hidden'' but important features are hard to discover, and even if we do manage to discover them, it is hard to instrument them. Moreover, such corporate decisions carry a huge cost associated with them-- to increase the marketing spend, the CEO will probably hire many more marketing professionals over the next few months. 


In reality, accurately instrumenting and modeling the generative process is hard. Despite this core challenge, what data scientists can do is provide transparency in terms of what they can and cannot conclude about the sales metric. The transparency comes in 2 forms- the first is in the form of ``model interpretability,'' where we have a way to explain why a model made a certain prediction. The other form of transparency is ``causality,'' where we want to ensure that our counterfactual is well understood, i.e., we can conclude that the final metric changes ``because'' of the change in the variable of interest, e.g., marketing spend, and not some other variable.    

In Section \ref{sec:review}, we first review the various available tools for model interpretability and causal inference or discovery, briefly explaining important concepts when needed. Section~\ref{sec:data} briefly describes a real-world dataset that will be used for our experiments in this paper, in accordance with the corresponding regulations. The input data has been transformed so that it is not possible to derive any sensitive information about the dataset. Details of how the features and labels were obtained cannot be shared, therefore we encode them as $F_1, F_2, \dots$. Section~\ref{sec:experiments} describes our findings and discusses the pros and cons of the various causality and model interpretability techniques.  Finally, we discuss conclusions and further extensions of our work in Section~\ref{sec:conclusion}.


\section{\label{sec:review}Review and Related Work}
\subsection{Model interpretability}

Among the techniques we discuss in this paper, model interpretability has been the most well studied. Model interpretabilty can be at 2 levels- local and global. Local interpretability is a feature-dependent prediction, where the feature importance for feature $i$, $\phi_i$, is a function of the value of the features $x$. For global interpretability measures, $\phi_i$ is independent of x. Whether to use local or global interpretability measures depends on the application-- if we need to provide explanations for individual predictions, for example, identifying the important features due to which the model recommended a product to a particular user. Global interpretability metrics are suitable in cases where we need to understand the overall behavior of the classifier, where we are concerned whether the model prediction is too biased on a single feature. In their book, \citet{Gaur2024-zk} provide an overview of the applications of model interpretability in e-commerce. There are also papers applying interpretability techniques to specific e-commerce use cases, such as recommenders \cite{Sahu2024}, product search \cite{xai-productsearch}, customer behavior \cite{azad-xai}, and customer satisfaction. \cite{ansari2023unlocking} 

Another classification of interpretability is into post-hoc interpretability and inherently interpretable modeling. Post-hoc interpretability is more commonly used, where we train a model that optimizes accuracy, and then invoke the necessary interpretability tools to explain model predictions. This allows us to be flexible in terms of the model used, because the explainability part is done ``post-hoc.'' The literature on inherently interpretable models is sparse, because it is difficult to design highly accurate ``white box models.'' This tradeoff between accuracy and interpretability has been discussed in the literature in detail.\cite{rane-tradeoff, dziugaite2020enforcing} Examples of models with high inherent interpretability are logistic regression and linear regression that have limited modeling power, whereas neural networks result in high accuracy but are extremely difficult to interpret. Moreover, interpretability measures can be model-dependent or model-agnostic. For example, Shapley Additive Explanations (SHAP) \cite{shap} and Locally Interpretable Model-agnostic Explanations (LIME) \cite{lime} are model agnostic, whereas methods such as integrated gradients and SmoothGrad only apply to models such as neural networks where the gradient with respect to model parameters is well-defined. In this paper, we focus on SHAP, because of its comprehensive capabilities and additive nature. SHAP values are grounded in cooperative game theory, and the Shapley value equation \cite{shapley:book1952} is the only function that satisfies the fairness axioms of linearity, efficiency, symmetry, and null values for non-contributing players in that cooperative game scenario. 

\subsection{Causal Inference}
While SHAP values are powerful and grounded firmly in theoretical principles, we cannot directly conclude that the feature with the highest Shapley value was the main cause behind the predicted sales for a given week, in the strategy planning example discussed in Section~\ref{sec:intro}. Confounding effects are common in complex, real-world use cases such as retail. SHAP values would exactly correspond to a causal effect if each of the features was an independent ``lever arm'' that the CEO could tune as she liked, while controlling every other possible lever arm at its original value. Such an ideal scenario is unlikely. Hence, we need a different framework to accurately model counterfactuals and confounders. Causal inference helps us achieve this objective. Several retail, manufacturing, and information technology organizations have succeeded in their objectives by using causal inference techniques. \cite{causalens} In retail and e-commerce, causal inference has been used in various use cases, such as recommender systems \cite{gao2023causal} and analyzing causal effects of interventions such as promotions \cite{AguilarPalacios2021}. Causal inference techniques have also been applied to analyze competition among retailers \cite{davis2022competitive}, impacts of an environmental regulation on secondary markets, and to what extent customers value the high-quality delivery service on a leading online retail platform \cite{li-phd}. 

Causal inference relies on various methods that estimate some causal effect, for example, average causal effect (ACE) and average treatment effect (ATE) \cite{causal-retail1, causal-retail2}. In this paper, we shall focus on double machine learning. Double machine learning \cite{double-ml} is a flexible tool for observational causal inference applications, such as retail sales prediction. It is called ``double'' machine learning because it uses the confounders to first train a model to deconfound the feature of interest, and then trains another machine learning model to estimate the average causal effect of changing that feature. In this case, this effect would be measured as a slope between the feature of interest and a target, e.g., sales. We should note that double machine learning is not suitable for cases with unobserved confounders. In such cases, we need to either create or exploit partial randomizations existent in the data that can ``break'' the correlation between the unobserved confounders and the feature(s) of interest. Depending on the application, some of the popular techniques in these cases are instrumental variables \cite{instrumental}, regression discontinuity \cite{regression-discontinuity}, and difference-in-difference \cite{did-causal}.   

\subsection{Causal discovery}
Causal inference models the complex interdependence between various features and their relations with the target variable of interest, e.g., sales. These can be modeled as a directed acyclic graph (DAG), and the principles of probabilistic graphical models can be applied to such graphs. However, the edges of these graphs and the directionality of these edges are often unknown. Causal discovery helps in this process. Learning this directed acyclic graph can be split into 2 problems-- a) learning the conditional probability distributions of the edges of a DAG, and b) given a set of samples, estimate a DAG that selects the appropriate causal edges between its nodes.

The first problem is less challenging and appears in many fields of machine learning, e.g., Baum-Welch algorithm for Hidden Markov Models. The second problem of structure estimation is challenging not only because of the usual challenges of causal inference but also because of the exponential complexity of the space of possible DAGs, which is super-exponential in the number of nodes. Popular methods are score-based structure estimation (BIC/BDeu/K2 scores, exhaustive search, and hill climb/tabu search), constraint-based structure estimation (e.g., PC algorithm), and hybrid structure estimation (e.g., MMHC algorithm \cite{Tsamardinos2006}). In the retail industry, such DAG estimation techniques have been used in recommender systems \cite{gao2024causal}, along with some other e-commerce or retail applications. The names of the features in the DAG could also be fed into a large language model to guess the most likely set of directed edges in the DAG. \cite{causal-biz-book} 





\section{\label{sec:data}Data}
The proprietary dataset used in this paper describes features and metrics for various product lines, from 2019 to 2024. Due to the proprietary nature of the dataset, we do not share individual values or summary statistics of the raw feature values. However, we share plots such as feature importance and correlation coefficients, because they do not reveal anything about the underlying data, and are sufficient for justifying the observations and conclusions we make in this paper. The findings in this paper generalize to other use cases in retail too.

The original dataset had 64 columns. In order to reduce the complexity of the problem, we focused on 7 features, named $F_1$ to $F_7$. These features were decided by recursively deleting features that have a correlation coefficient of more than 0.3 with at least one of the other features. $Y$ is the variable of interest, and $F_1, \dots, F_7$ are the shortlisted features, which denote some generic retail business metrics.

\begin{figure}[!htbp]
\centering
\begin{subfigure}{\linewidth}
  \centering
  \includegraphics[width=.8\linewidth]{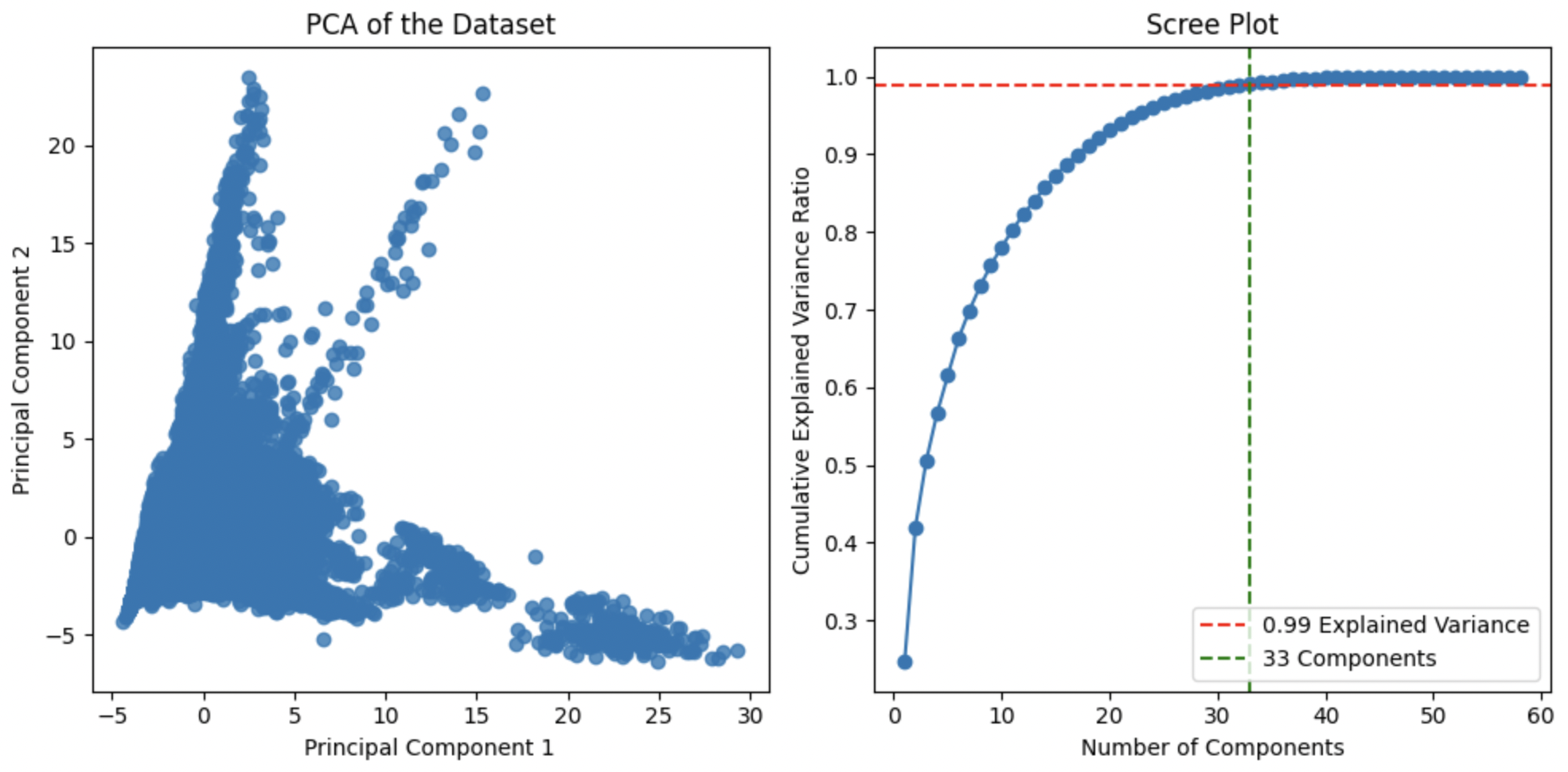}
  \caption{  \label{fig:pca}Principal components of the raw dataset.}
\end{subfigure}

\begin{subfigure}{\linewidth}
  \centering
  \includegraphics[width=.5\linewidth]{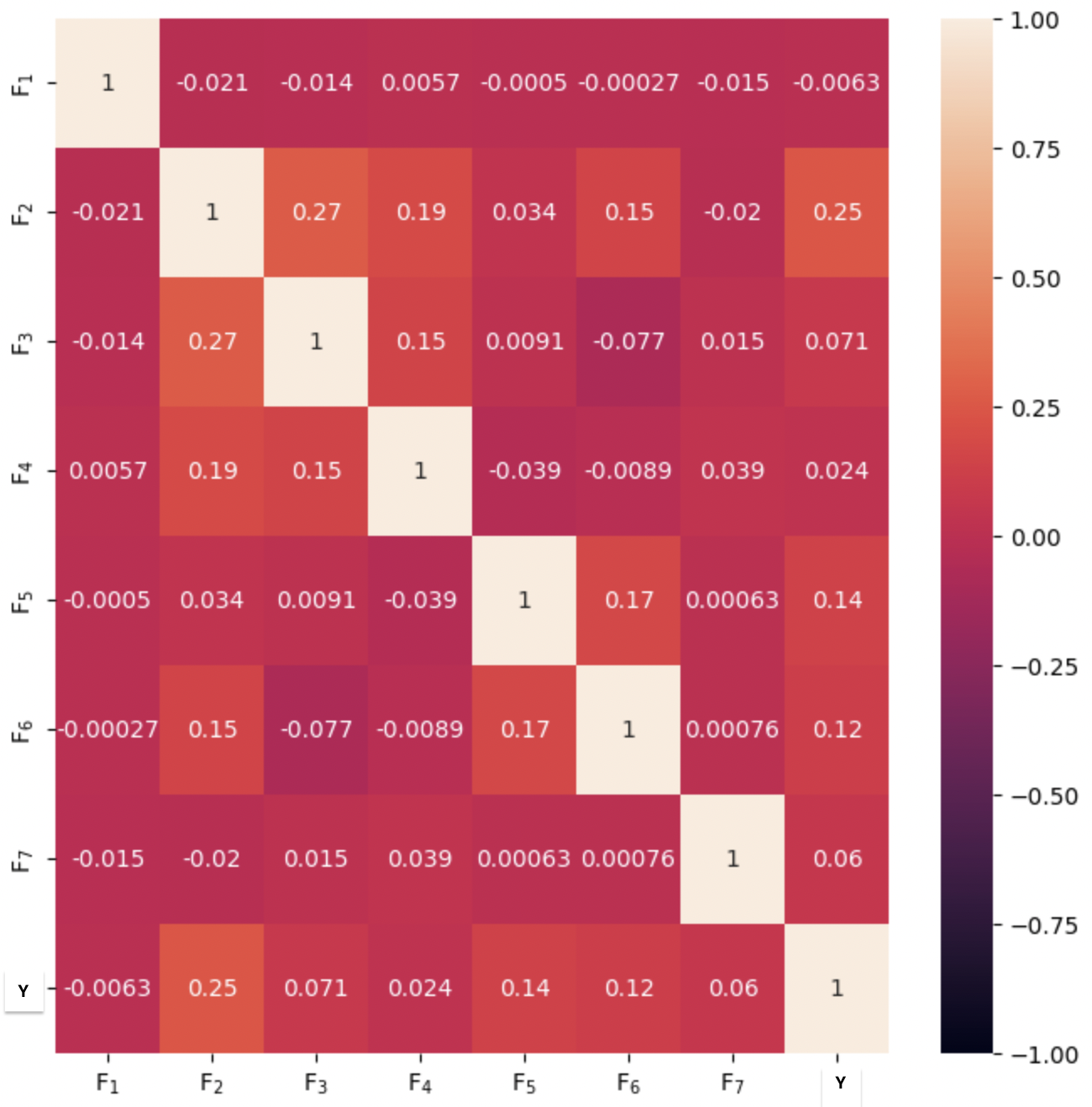}
  \caption{\label{fig:correl}Correlation coefficients among the shortlisted features $F_1, \dots, F_7$ and the target metric $Y$.}
\end{subfigure}
\caption{Summary plots of the dataset.}
\end{figure}

While this reduction drastically simplifies modeling, the data might higher dimensional than the selected number of features $F_1, \dots, F_7$. The delayed onset of the plateau in the Scree plot in Fig.~\ref{fig:pca} supports this possibility. 

\section{\label{sec:experiments}Experiments}
\subsection{Model interpretability}
We train multiple tree-based regression models between the features $F_{1}, \dots F_7$ and $Y$, as shown in Table~\ref{tab:errors}. We report loss as the mean absolute error divided by the average of the target $Y$, in order to create a dimensionless loss metric, namely the mean scaled absolute error (MSAE). In other words,
\begin{equation}
    MSAE = \frac{\mathrm{mean}(\|Y_{true}-Y_{pred}\|)}{\mathrm{mean}(Y_{true})}
\end{equation}

We used hyperparameter optimization to help choose the best-performing hyperparameters for each of the 3 models.

Although the explainable boosting regressor has a worse MSAE when compared to the other, more ``black-box'' models, the SHAP values for the inherently explainable regressor have less variance. This is an interesting finding, which might be attributed to the inherent explainability in these models. Moreover, the qualitative features of the SHAP scatter plots do not change significantly across the 3 models. Evaluating explainability methods is an active area of research \cite{Hsieh2019EvaluationsAM, sanity-saliency}. In addition to these metrics, it is also important to have other metrics that are more aligned with the business needs. For example, we could survey company leaders to vote on which feature they think was the most important for a given week. Based on their feedback, we could create a ground truth dataset for feature importance values. Each model and explainability technique could be benchmarked against this ground truth dataset.

\begin{figure}[!htbp]
  \centering
  \includegraphics[width=9cm]{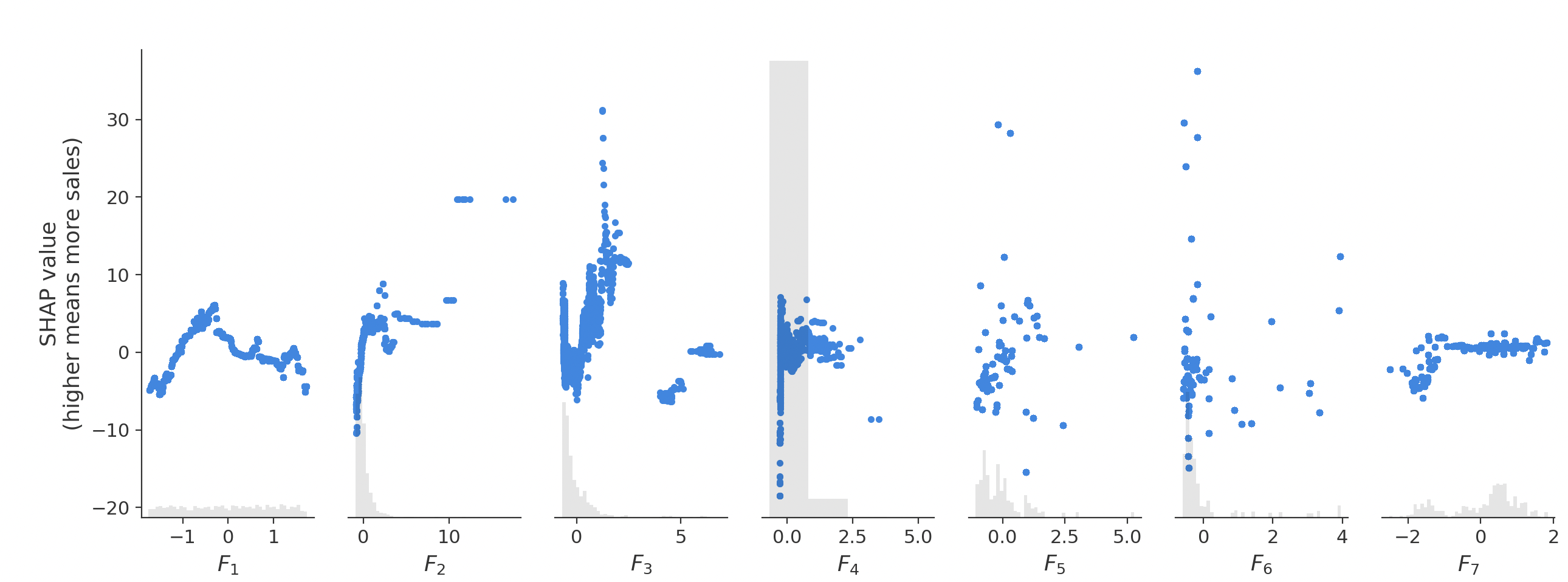}
  \caption{\label{fig:shap_ebm}SHAP values for the Explainable Gradient Boosting Regressor.}
\end{figure}

\begin{figure}[!htbp]
  \centering
  \includegraphics[width=9cm]{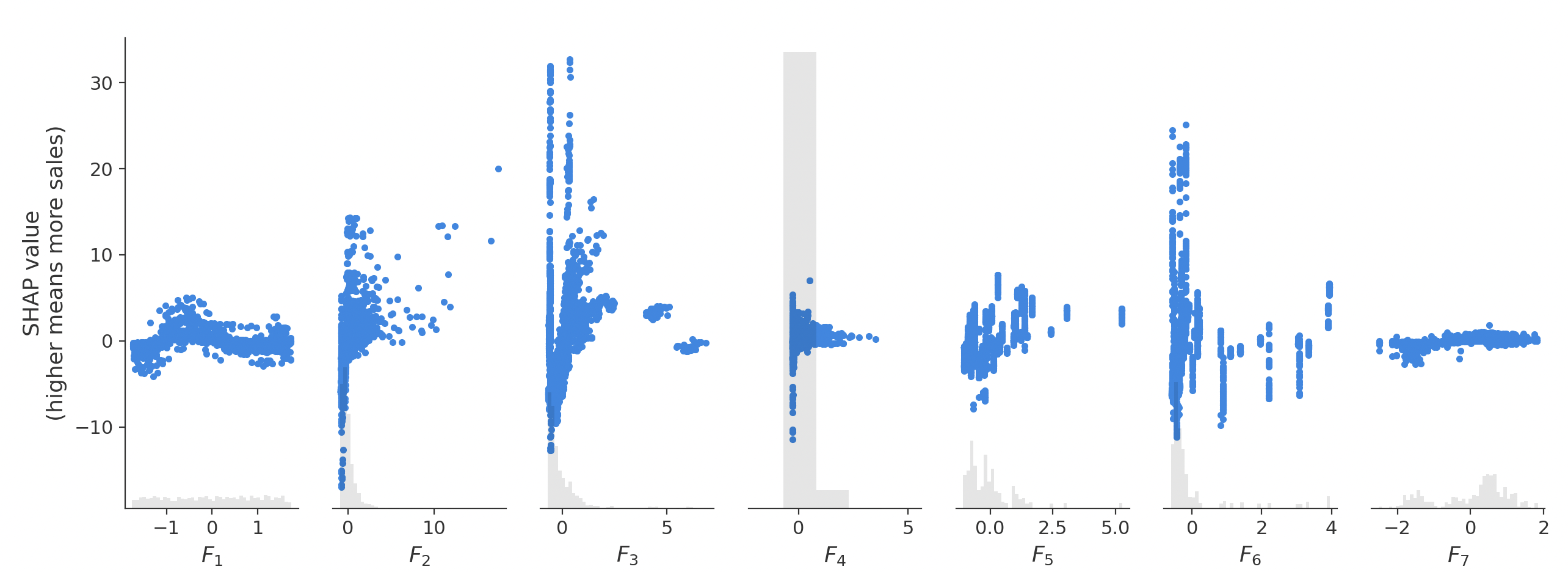}
  \caption{ \label{fig:shap_rf}SHAP values for the Random Forest Regressor.}
\end{figure}

\begin{figure}[!htbp]
  \centering
  \includegraphics[width=9cm]{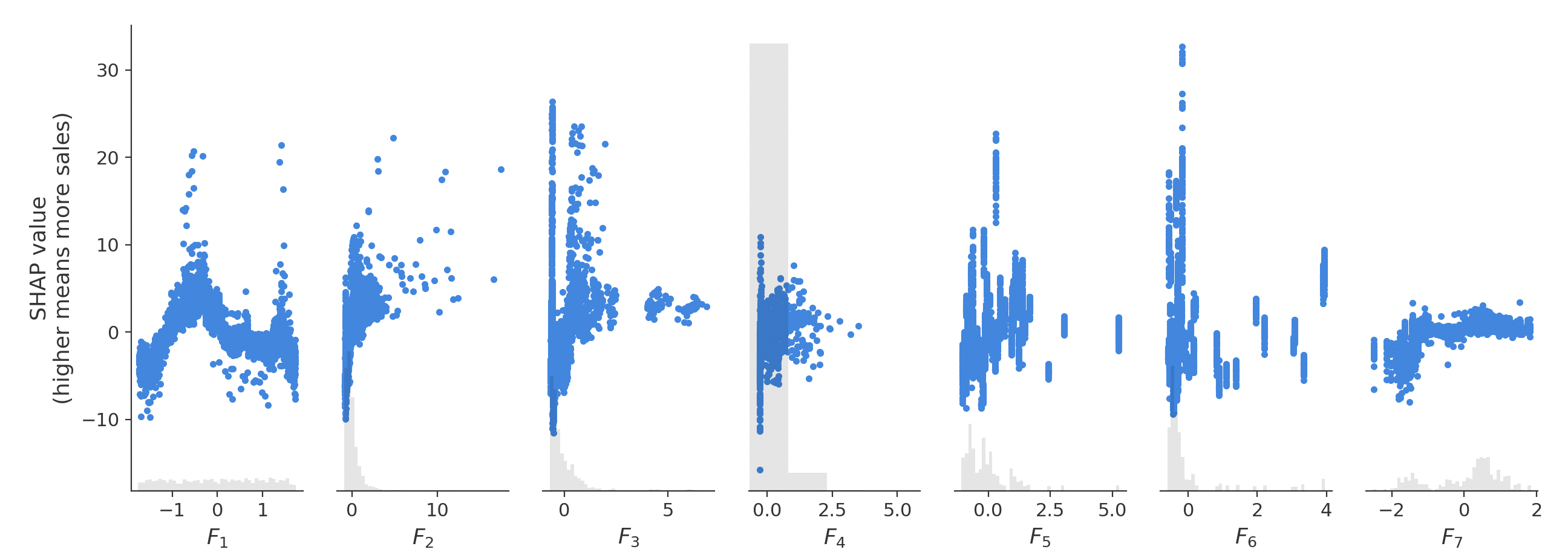}
  \caption{\label{fig:shap_xgboost}SHAP values for the XGBoost Regressor.}
\end{figure}

We briefly discuss the SHAP value plots in Figs.~\ref{fig:shap_ebm}, \ref{fig:shap_rf}, and \ref{fig:shap_xgboost} for the features $F_1$ to $F_7$.  SHAP values for $F_1$ (seasonality) show a rise and fall, which is expected because sales increase and decrease based on seasonality features such as day/month of the year, distance to the nearest holiday, etc. SHAP values for $F_2$ (underperforming products) increase, plateau, and then decrease slightly as $F_2$ increases. This behavior is counterintuitive, because increasing underperforming products should have a monotonically decreasing effect on the sales. Increase in $F_3$ (inventory) first leads to higher contributions to sales, and then leads to a decline for much large values. This might be because having too much inventory might lead to other issues that preclude sales. 
Contribution to sales should monotonically decrease as $F_4$ (out-of-stock items) increase, but the reverse behavior is seen when $F_4 \rightarrow 0$. $F_5$ (negative customer perception) should lead to lower sales, but the SHAP plot does not show a clear dependence. On the contrary, we see a small upward trend. There is an initial increase in SHAP values as a function of $F_7$ (stock market trends), but then the SHAP values saturate, similar to other features such as $F_2$ and $F_4$. In later subsections, we try to understand why this is so. 

\begin{table}
  \begin{tabular}{p{0.2\linewidth}p{0.4\linewidth}p{0.15\linewidth}p{0.15\linewidth}}
    \toprule
    Model name & Hyperparameters (others set to default) & Train set MSAE & Validation set MSAE\\
    \midrule
    XGBoost & \verb|n_estimators=175|, \verb|colsample_bytree=0.8|, \verb|colsample_bylevel=0.8|, \verb|eta=0.1| & 0.106 & 0.135\\
    \\
    Random Forest & \verb|n_estimators=100|, \verb|max_depth=7| & 0.221 & 0.234\\
    \\
    Explainable Gradient Boosting & \verb|interactions=1| & 0.211 & 0.236\\
  \bottomrule
\end{tabular}
  \caption{\label{tab:errors}Comparing model performances. Note that each model had an earlier preprocessing step in which the values were scaled using the StandardScaler in scikit-learn.}
\end{table}

\subsection{Causal inference}
\begin{figure}[!htbp]
\centering
\begin{subfigure}{\linewidth}
  \centering
  \includegraphics[width=.8\linewidth]{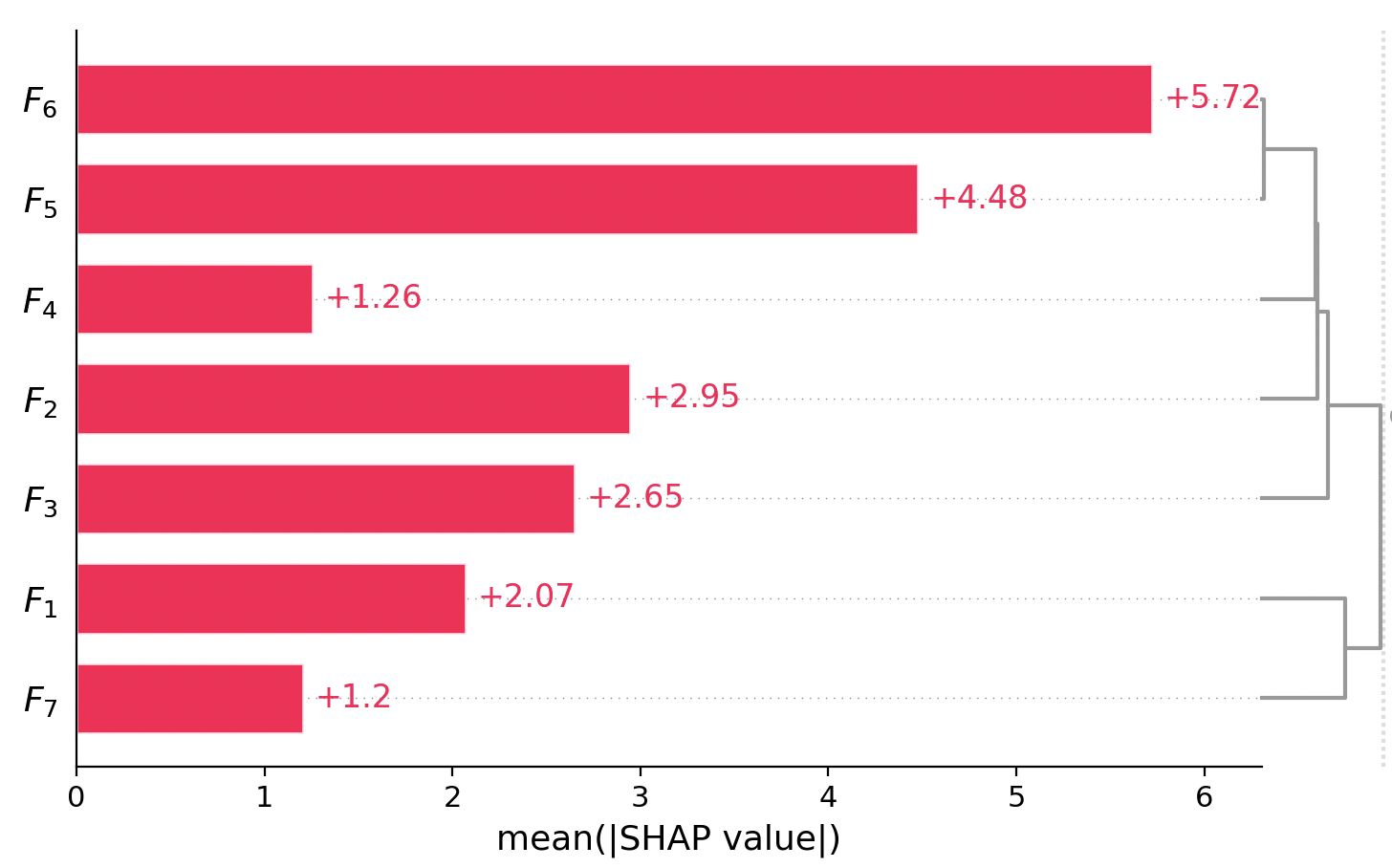}
  \caption{  \label{fig:hclust}SHAP-based global feature importance for the Explainable Gradient Boosting Regressor in Fig.~\ref{fig:shap_ebm}. We also provide a redundancy clustering dendogram on the right side of the bar chart.}
\end{subfigure}

\begin{subfigure}{\linewidth}
  \centering
  \includegraphics[width=.8\linewidth]{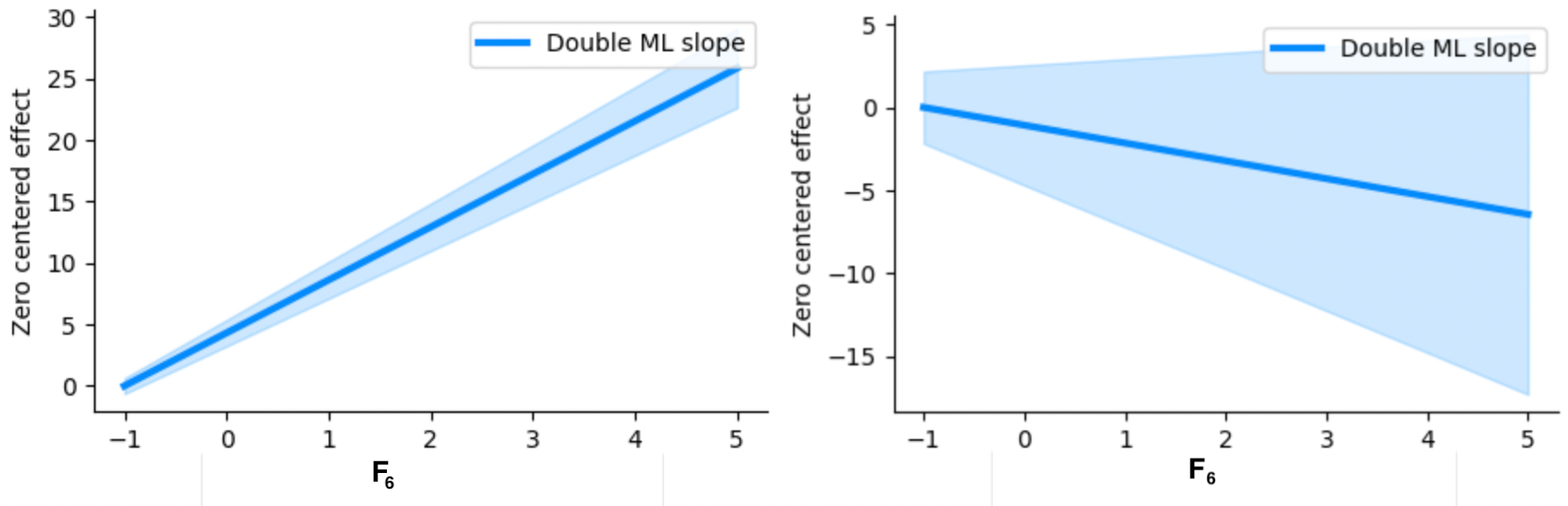}
  \caption{\label{fig:dml}Double machine learning predictions for the average causal effect of the seller unresponsiveness variable $F_6$. By controlling just $F_5$, we get the wrong sign of the causal effect (+ve slope instead of the expected -ve slope) in the left plot. In the right plot, we get the correct behavior by controlling $F_2$, $F_4$, and $F_5$.}
\end{subfigure}
\caption{Global importance, redundancy clustering, and average causal effect of seller unresponsiveness on sales using double ML.}
\end{figure}

For usual machine learning models to deliver causal results, the features need to be independent not only of other features in the model, but also of unobserved confounders. In order to understand the redundancies better, we perform single-link hierarchical clustering on the features, as shown in Fig.~\ref{fig:hclust} next to the global feature importance plot. For typical tabular data such as ours, this kind of clustering results in much more accurate measures of feature redundancy than unsupervised methods such as correlation \cite{feature-cluster-shap}. 


From the global feature importance bar chart in Fig.~\ref{fig:hclust}, it seems that $F_6$, $F_5$, $F_2$ are the most important features. However, from the clustering, it seems that there is redundancy between $F_5$ (negative customer perception) and $F_6$ (lack of seller responsiveness), and they have some redundancy with $F_4$ (out-of-stock) and $F_2$ (underperforming products) as well. Such (observable) confounding can create misleading SHAP values for our Explainable Boosting Regressor, because a predictive model unaware of the underlying causal structure might choose one of the several redundant features, rather than choosing the feature that has no parents. In order to test whether this is true, we can control $F_2$, $F_4$, and $F_5$ successively and see whether there is a significant change in the estimated causal effect of $F_5$ on $Y$ (sales), when compared to just controlling for $F_5$. This is possible through double ML. Controlling for more confounding variables leads to the correct sign of causal effect, i.e., higher the unresponsiveness $F_6$, lower the sales $Y$.



\section{\label{sec:conclusion}Conclusion and Further Work}
In this paper, we reviewed model explainability and causal inference methods that can be applied to retail datasets. We found that the inherently explainable model has lower variance on SHAP values despite its lower accuracy when compared to a normal XGBoost classifier. We then explored double machine learning and discovered that the average causal effect changed sign when controlling for possible confounders, and that Shapley values are not reliable when it comes to discovering causal relationships. Further work could involve the integration of SHAP values into causal inference, for example, c-SHAP \cite{heskes2020causal}, and expanding to more features while using computationally efficient methods for SHAP values and DAG estimation.  

\newpage
\bibliographystyle{plainnat}
\bibliography{custom}

\end{document}